\documentclass[10pt,twocolumn,letterpaper]{article}

\usepackage[pagenumbers]{wacv} % To force page numbers, e.g. for an arXiv version

\usepackage{graphicx}
\usepackage{amsmath}
\usepackage{amssymb}
\usepackage{booktabs}
\usepackage{times}
\usepackage{epsfig}
\usepackage{graphicx}
\usepackage{amsmath}
\usepackage{amssymb}
\usepackage{multirow}
\usepackage{bm}
\usepackage[bottom]{footmisc}
\usepackage{colortbl}

\usepackage{adjustbox}
\usepackage{setspace}
\usepackage[normalem]{ulem}

\usepackage{booktabs}
\usepackage{subcaption}
\usepackage{parselines}
\usepackage{lineno}
\usepackage{setspace}
\usepackage{emptypage}
\usepackage{rotating}
\usepackage{booktabs}
\usepackage{makecell}
\usepackage{algorithm,tabularx}
\usepackage{csquotes}
\usepackage{xhfill}
\usepackage{xcolor}
\usepackage{soul}
\usepackage{bbm}
\usepackage{dsfont}
\usepackage{float}
\usepackage[cal=cm]{mathalpha}
\usepackage[pagebackref,breaklinks,colorlinks]{hyperref}

\usepackage[capitalize]{cleveref}
\crefname{section}{Sec.}{Secs.}
\Crefname{section}{Section}{Sections}
\Crefname{table}{Table}{Tables}
\crefname{table}{Tab.}{Tabs.}

\def\wacvPaperID{2486} % *** Enter the WACV Paper ID here
\def\confName{WACV}
\def\confYear{2025}

\begin{document}
\raggedbottom
%%%%%%%%% TITLE - PLEASE UPDATE
\title{ROADS: Robust Prompt-driven Multi-Class Anomaly Detection \\under Domain Shift}

\author{Hossein Kashiani\qquad Niloufar Alipour Talemi\qquad Fatemeh Afghah\\
Clemson University\\
{\tt\small \{hkashia, nalipou, fafghah\}@clemson.edu
}
}

% \author{Hossein Kashiani\\
% Clemson University\\
% Institution1 address\\
% {\tt\small hkashia@clemson.edu}
% % For a paper whose authors are all at the same institution,
% % omit the following lines up until the closing ``}''.
% % Additional authors and addresses can be added with ``\and'',
% % just like the second author.
% % To save space, use either the email address or home page, not both
% \and
% Niloufar Alipour Talemi\\
% Institution2\\
% First line of institution2 address\\
% {\tt\small secondauthor@i2.org}
% }
\maketitle

%%%%%%%%% ABSTRACT
\begin{abstract}

Recent advancements in anomaly detection have shifted focus towards Multi-class Unified Anomaly Detection (MUAD), offering more scalable and practical alternatives compared to traditional one-class-one-model approaches. However, existing MUAD methods often suffer from inter-class interference and are highly susceptible to domain shifts, leading to substantial performance degradation in real-world applications. In this paper, we propose a novel robust prompt-driven MUAD framework, called ROADS, to address these challenges. ROADS employs a hierarchical class-aware prompt integration mechanism that dynamically encodes class-specific information into our anomaly detector to mitigate interference among anomaly classes. Additionally, ROADS incorporates a domain adapter to enhance robustness against domain shifts by learning domain-invariant representations. Extensive experiments on MVTec-AD and VISA datasets demonstrate that ROADS surpasses state-of-the-art methods in both anomaly detection and localization, with notable improvements in out-of-distribution settings.
\end{abstract}

\section{Introduction}
\label{sec:intro}

Anomaly detection (AD) is a critical task across domains like industrial defect detection \cite{bergmann2019mvtec}, medical imaging \cite{fernando2021deep}, and surveillance \cite{xia2020synthesize}, focusing on identifying data instances that deviate from the expected norm. Due to scarcity and high cost of obtaining labeled anomaly data, unsupervised approaches are predominantly employed for anomaly detection \cite{liu2023simplenet, zavrtanik2021draem, deng2022anomaly}. These unsupervised methods typically learn patterns from a set of normal training samples to detect anomalies in test data.

Conventional AD methods \cite{liu2023simplenet, zavrtanik2021draem, deng2022anomaly,tien2023revisiting, zhang2023prototypical} rely heavily on the one-class-one-model paradigm, where a separate model is trained for each class of objects. This paradigm quickly becomes impractical in real-world, large-scale applications, as it necessitates considerable memory and computational resources. Furthermore, these methods generally assume that training and test data are drawn from the same distribution \cite{deng2022anomaly,yang2023ad}. In practice, this assumption is often violated due to factors such as changes in lighting conditions, object poses, or environmental factors, leading to significant distribution shifts. Such shifts can drastically degrade the performance of AD systems that are optimized for distributional similarity between training and test data.

\begin{figure}[t] 
\centering
    \includegraphics[scale = 0.33]{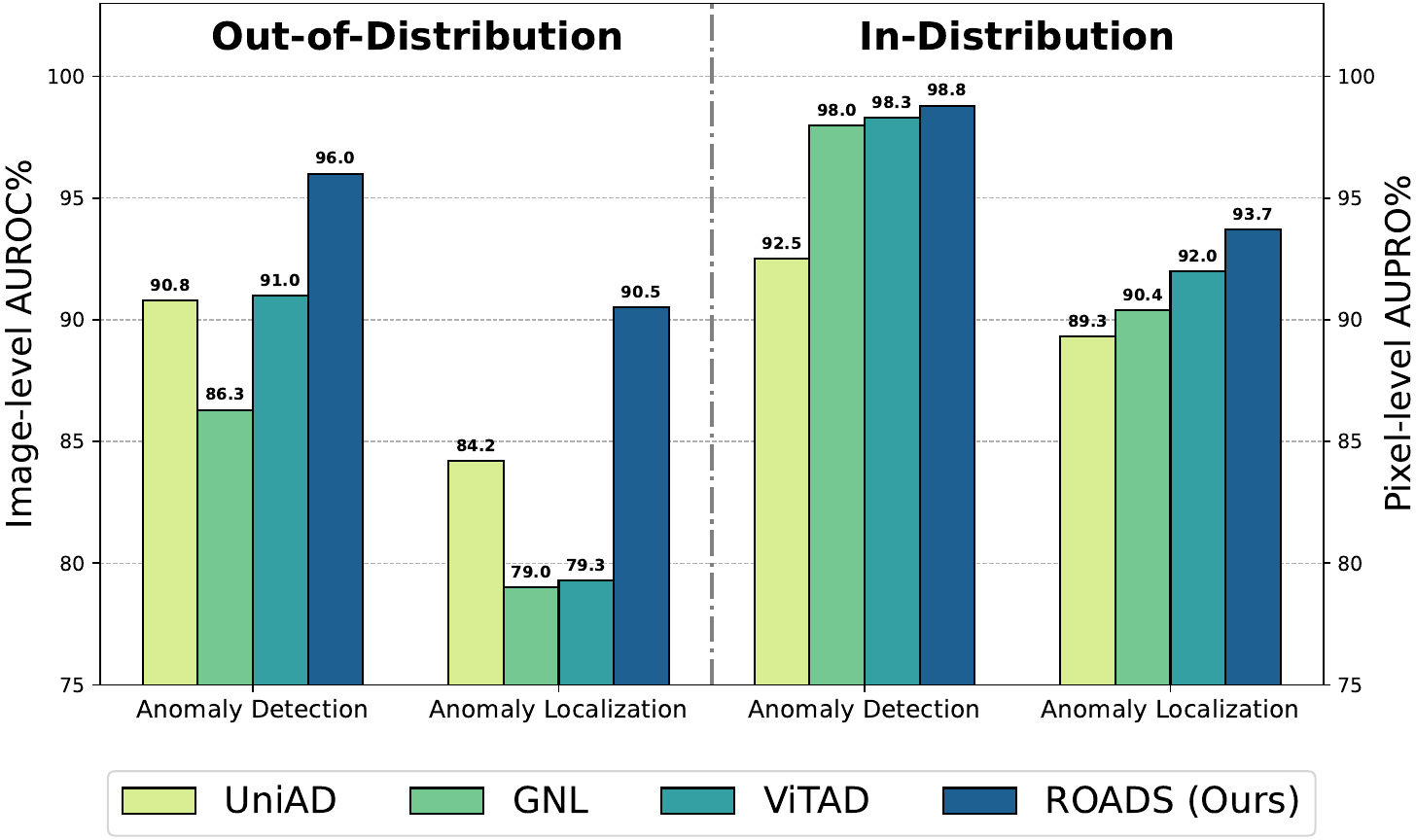}%width=0.5\textwidth mask5 not good 
    \caption{Comparison of multi-class anomaly detection methods on the MVTec-AD \cite{bergmann2019mvtec} in both in-distribution and out-of-distribution settings. Image-level AUROC and pixel-level AUPRO are used for anomaly detection and localization evaluations. Methods, including UniAD \cite{Unified2024}, GNL \cite{cao2023anomaly}, and ViTAD \cite{vitad} , are evaluated under a unified setup. Our method exhibits exceptional robustness in out-of-distribution scenarios while outperforming competing approaches in in-distribution settings.}
    \label{highlight}
\end{figure}

Recent studies have sought to address the limitations of traditional anomaly detection by focusing on multi-class unified anomaly detection (MUAD) \cite{mambaad,invad,vitad,neurips2023hvq,Unified2024,zhao2023omnial}. These approaches aim to unify multiple classes within a single framework to overcome the challenges posed by large-scale datasets. However, despite advancements in MUAD, existing methods still struggle to maintain robustness under distribution shifts—a critical challenge in real-world applications. As illustrated in Figure \ref{highlight}, state-of-the-art (SOTA) studies experience notable performance degradation when evaluated on out-of-distribution (OOD) data. This vulnerability arises because many MUAD studies rely on architectures tailored to operate under the assumption of consistent data distributions, limiting their ability to generalize to OOD scenarios.

Moreover, current MUAD methods \cite{Unified2024, zhao2023omnial} typically handle all classes within a shared framework, leading to a closely intertwined semantic space among different classes. This entanglement increases the risk of catastrophic forgetting, where models lose previously learned class-specific features when new classes are introduced and struggle to establish distinct boundaries between classes. This issue is exacerbated under OOD conditions, where domain shifts—such as changes in lighting or object poses—further obscure the distinction between classes, leading to frequent misclassifications of normal samples as anomalies. These challenges highlight the necessity for robust, unified models capable of handling MUAD under diverse distribution shifts.

Recently, prompting has gained significant prominence in language-visual multi-modal systems, such as CLIP \cite{yu2023visual}. These frameworks dynamically regulate models for specified tasks based on provided prompt information. Motivated by this, and to address the aforementioned challenges, we propose a novel prompt-driven teacher-student framework tailored for MUAD in the presence of domain shifts. 
Our approach introduces a class-aware prompt integration mechanism that leverages class-specific prompts to encode critical information about anomaly classes, guiding our detector in effectively distinguishing between different anomaly classes by separating their semantic spaces. These prompts act as auxiliary inputs, offering precise contextual information that pushes boundaries of different classes farther apart. With the integration of class-specific prompts, our framework preserves the features of previously learned classes while minimizing interference from new classes, enhancing the overall detection performance.

Additionally, to enhance the robustness of our model against distribution shifts, we propose a domain adapter to dynamically adjust our anomaly detector under domain shifts. As affine transformation parameters in Adaptive Instance Normalization (AdaIN) represents image styles \cite{dumoulin2017a,huang2017arbitrary}, the domain adapter processes input images into domain-invariant style codes for the AdaIN layers to account for domain variations. The domain adapter learns domain-invariant representations by enforcing style consistency regularization between in-distribution (ID) source domain and OOD target domain samples. By learning domain-invariant representations through domain style consistency learning, we reduce the divergence of style codes from different domains.

During inference, the model adapts to each test image individually, using the domain adapter to generate domain-invariant style codes that adjust feature representations in real time. This adaptive approach ensures our anomaly detection model can generalize effectively under distribution shifts. Our method demonstrates superior performance across both ID and OOD scenarios, as shown in Figure \ref{highlight}. In summary, the main contributions of this paper are threefold:

\begin{itemize}
\item We propose a robust prompt-driven MUAD framework that features a novel class-aware prompt integration mechanism. This mechanism dynamically learns and incorporates class-specific prompts, effectively reducing inter-class interference and improving both anomaly detection and localization.

\item We introduce a domain adapter to align the styles of OOD target domains with those of the ID source domain, ensuring consistent feature representation across varying domains.

\item The proposed method demonstrates significant improvements over SOTA approaches across challenging MVTec-AD \cite{bergmann2019mvtec} and VISA \cite{bergmann2019mvtec} benchmarks, with extensive evaluations confirming its superior robustness in handling distribution shifts.

\end{itemize}

\section{Related Works}
\label{sec:Related}
\noindent\textbf{Unsupervised Anomaly Detection}
Conventional unsupervised anomaly detection studies \cite{deng2022anomaly,tien2023revisiting, zhang2023prototypical} typically build separate models for each object class, which can be impractical for large-scale applications due to high memory consumption and computational costs. Mainstream approaches in this domain include embedding-based methods and reconstruction-based methods \cite{cao2024survey}. Embedding-based methods employ networks pre-trained on ImageNet \cite{deng2009imagenet} to map
normal features to a compact space and identify anomalies by differentiating between normal and anomalous data \cite{roth2022towards,rudolph2021same}. These methods have shown competitive performance in single-class settings but face scalability issues as the number of categories increases. On the other hand, reconstruction-based methods \cite{ristea2022self,liang2023omni,roth2022towards,akcay2019ganomaly,perera2019ocgan} often rely on self-training an encoder-decoder framework to reconstruct input images. Reverse distillation (RD) \cite{tien2023revisiting,deng2022anomaly,guo2024recontrast} is a line of studies in reconstruction category that employs a student decoder to restore the multi-scale representations of a pre-trained teacher encoder from one-class embeddings. These studies assume that a student decoder trained solely on normal samples could replicate the features of normal regions.

 \begin{figure*}[t] 
\centering
% \hspace{2cm} % Add a gap of 1 inch on the left side
\includegraphics[width=0.98\textwidth]{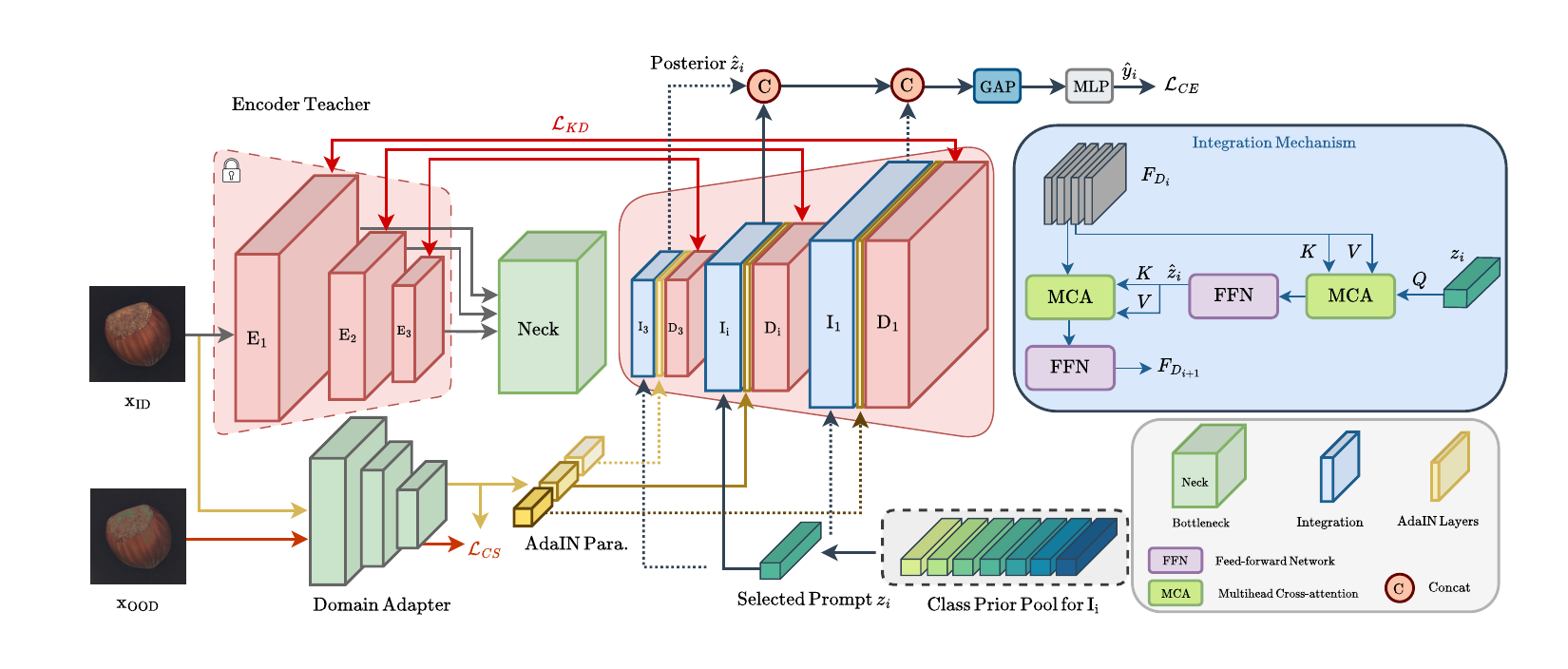}
\caption{Overview of the proposed ROADS framework. First, the domain adapter aligns  the features
of the OOD target domain. Then, the corresponding learned class-specific prompt tokens are selected from the pool and integrated into the student decoder.}
\label{framework}
\end{figure*}

\noindent\textbf{Multi-class Unsupervised Anomaly Detection.} To improve the efficiency and scalability of anomaly detection models, recent studies \cite{ader, mambaad,he2024diffusion,guo2024dinomaly, yao2024hierarchical} have introduced the concept of using unified models in MUAD to handle multiple data sources. This approach is memory-efficient, as it avoids the need to train separate models for each class. For instance, UniAD \cite{Unified2024} tackled the shortcut problem by modifying the transformer network and introducing a hierarchical query decoder, neighbor masking attention module, and feature jittering strategy. 
 Similarly, OmniAL \cite{zhao2023omnial} proposed using panel-guided synthetic anomaly data for training, and HVQ-Trans \cite{neurips2023hvq} utilized a hierarchical codebook mechanism to prevent shortcuts. Also, ViTAD \cite{vitad} proposed a symmetrical vision transformer with global modeling for MUAD. However, these methods \cite{Unified2024,neurips2023hvq,zhao2023omnial,vitad,guo2024dinomaly} often fail to consider the unique aspects of MUAD in terms of different categories. In real-world applications, identifying anomaly classes across various data sources is relatively straightforward, such as categorizing different products on distinct production lines. It is logical to enhance existing MUAD models by leveraging the information provided by these anomaly classes.

\noindent\textbf{Domain Generalization.}
Machine learning algorithms often assume that training and test data are independent and identically distributed (i.i.d.), but this assumption rarely holds in real-world scenarios, leading to performance drops. Domain generalization aims to overcome this by learning domain-invariant representations using multiple or single-source domains \cite{wang2022generalizing}. Domain generalization is crucial for anomaly detection, especially when training and test data differ significantly. Traditional AD methods \cite{ader, mambaad,he2024diffusion,tien2023revisiting,deng2022anomaly,guo2024recontrast} assume the same distribution for training and test data, which is rarely practical, causing poor performance under domain shifts. Few studies have tackled domain shift in anomaly detection, with most approaches either focusing on one-class detection \cite{yang2023ad} or requiring target data for domain adaptation \cite{deng2022anomaly}, limiting their applicability in multi-class scenarios. Early work by Cao et al. \cite{yang2023ad} proposed to learn domain-invariant features for a single-class anomaly detection model with test time adaptation. However, GRL \cite{yang2023ad} is computationally expensive, less effective in multi-class settings, and relies on source data during test-time adaptation, which limits its practicality. Our approach addresses these challenges by training a domain adapter that dynamically adjusts the detector to diverse distribution shifts on the fly.

\section{Methodology}
\label{sec:Methodology}

\subsection{Problem Definition}

Let \( \mathcal{S} \) and \( \mathcal{T} \) represent the ID source and OOD target domains in different classes. The source domain \( \mathcal{S} \) is used for both training and testing, while the target domain \( \mathcal{T} \) is only used during inference.  Only normal data from \( \mathcal{S}\) is available for training, denoted as \( \mathcal{D}_s = \{x \in  \mathcal{S}  \mid y = 0\} \), where \( y \in \{0, 1\} \) indicates whether a sample \( x \) is normal (\( y = 0 \)) or abnormal (\( y = 1 \)). During testing, data can be normal or abnormal and may come from either \( \mathcal{S} \) or \( \mathcal{T} \), represented as \( \mathcal{D}_t = \{x \in S \cup \mathcal{T} \mid y \in \{0, 1\}\} \). The objective of robust MUAD is to detect anomalies in different classes in \( \mathcal{D}_t \) regardless of whether they originate from \( \mathcal{S} \) or \( \mathcal{T} \).

\subsection{Reverse Distillation Framework}

RD serves as a foundational technique for unsupervised anomaly detection. In this framework, a teacher-student model utilizes an encoder-decoder architecture where knowledge is transferred from the deeper layers of the teacher encoder $E$ to the earlier layers of the student decoder $D$, meaning high-level, semantic information is passed on first. By employing multi-scale feature-based distillation, the student decoder $D$ tries to replicate the functionality of the teacher encoder $E$. The rationale behind the multi-scale feature-based distillation is that the shallow layers of a neural network focus on extracting local features associated with low-level details (such as color, edges, and texture), whereas the deeper layers, with their larger receptive fields, capture more global semantic and structural information. Therefore, discrepancies between the low- and high-level features in the teacher-student model can indicate local and broader structural anomalies.

% \xi \zeta  \phi \varphi \Phi 	\psi \Psi \varphi

Our RD framework operates with five core components: a pre-trained teacher encoder $E$, a trainable bottleneck module $\varphi$, hierarchical class-aware prompt integration mechanism, a domain adapter $\xi$, and a student decoder $D$. To construct the teacher encoder and bottleneck module, we adhere to the baseline RD framework \cite{guo2024recontrast,tien2023revisiting,deng2022anomaly}. Specifically, the teacher encoder, which has been pre-trained on ImageNet \cite{deng2009imagenet}, extracts multi-scale spatial features \( F_{E_i}  = E_i(x)\) from the given input image $ x \in \mathbb{R}^{C \times H \times W} $ drawn from the normal training set 
$ \mathcal{D}_s$, where $C$, $H$,
and $W$ is the channel, height, and width of $x$ and $i$ is the layer  index in \(\{F_{E_i}\}_{i=1}^{M} \in \mathbb{R}^{C_i \times H_i \times W_i}\). The bottleneck module $\varphi$ then compresses\(\{F_{E_i}\}_{i=1}^{M}\) to minimize redundancy and generate an intermediate representation \( \phi \). Subsequently, the student decoder $D$ reconstructs the compressed features \( \phi \) as \( F_{D_i}  = D_i(\phi)\). Unlike a symmetrical autoencoder, the student decoder $D$ mirrors the architecture of teacher encoder but replaces down-sampling operations with up-sampling ones to reconstruct the feature maps of the teacher as \(\{F_{D_i}\}_{i=1}^{M} \in \mathbb{R}^{C_i \times H_i \times W_i}\). This reconstruction process aims to align the feature-level representations of the teacher and student networks.

\subsection{Hierarchical Class-Aware Prompt Integration}

Anomaly detection in a multi-class setting is particularly challenging due to the inter-class interference between different anomaly classes. To overcome this issue, we introduce a hierarchical integration mechanism that dynamically learns and incorporates class-specific prompt tokens directly from diverse anomaly classes into our anomaly detector. This class-aware approach guides the student decoder in reconstructing the feature representations generated by the teacher encoder, with the class-specific prompt tokens serving as constraints throughout the reconstruction process. This strategy effectively mitigates inter-class interference, enhancing the ability of our framework to detect anomalies across multiple classes.

            For each class in our dataset, a specific class prior is encoded as a learnable prompt token. We initialize a class prior pool \( \bm{Z} \in \mathbb{R}^{N \times l \times M_t} \) with Xavier initialization \cite{glorot2010understanding}, where \( l \) denotes the token length, \( M_t \) is the dimension of each token, and \( N \) represents the number of classes. Each class in the dataset is associated with multiple prompt tokens that encapsulate the prior information specific to that class. When an image $x$ with a particular class \( i \) is processed, the corresponding class-specific prompt tokens \( \bm{z}_{i} \in \mathbb{R}^{l \times M_t} \) is selected from the pool \( \bm{Z} \) via an anomaly classifier $\zeta(x)$. This token is then used by the student decoder $D$ to reconstruct the features from the intermediate representation \( \phi \). This design enables the decoder to adaptively select and incorporate class-specific prior information, effectively mitigating inter-class interference by embedding precise contextual information for each anomaly class.

The integration mechanism fuses the knowledge encapsulated in these prompt tokens with the feature representations of the student decoder using a cross-attention module. By treating prompt tokens as Query vector and the feature representations as Key and Value vectors, this module facilitates efficient and global information exchange between the class-specific prompt tokens and the feature representations of the student decoder while maintaining computational efficiency. Given the reshaped intermediate representation \( {F_{D_i}} \in \mathbb{R}^{n \times M_t} \) (where \( n = H \times W \)) and the prompt tokens \( \bm{z}_i \in \mathbb{R}^{l \times M_t} \), the mechanism first aggregates image-specific information from the student decoder to generate posterior tokens \( {\bm{\hat{z}}_{i}} \). 
\begin{equation}\label{e1}
    {\bm{\hat{z}}_{i}} = \operatorname{FFN}\Biggl(\operatorname{{L}}\Bigl(\underbrace{\operatorname{MCA}\big(\text{L}(\bm{z}_i)), \operatorname{L}(F_{D_i})\big)}_{\operatorname{Cross-Attention} \operatorname{Output}} + \bm{z}_i\Bigr)\Biggr) + \bm{z}_i,
\end{equation}
\noindent where \(\operatorname{FFN}\) is a feed-forward network, \( \operatorname{MCA} \) denotes multi-head cross-attention, and \( \operatorname{L} \) represents layer normalization. The $\operatorname{MCA}$ module in Eq. \ref{e1} treats the prompt tokens \( \bm{z}_i \) as the Query vector \( Q \), and also treats the feature representation \( F_{D_i} \) as the Key vector \( K \) and Value vector \( V \). The $\operatorname{MCA}$ module computes the attention mechanism as follows:
\begin{equation}
    \operatorname{MCA}(Q, K, V) = \operatorname{Softmax}\left(\frac{Q K^\top}{\sqrt{d_k}}\right) V,
\end{equation}
\noindent where \( Q = \bm{z}_i W_Q \), \( K = F_{D_i} W_K \), \( V = F_{D_i} W_V \), and \( d_k \) is the dimension of the Key vectors. The outputs from multiple attention heads are then concatenated and projected to form the final cross-attention output. Subsequently, these posterior tokens are reintegrated into $F_{D_i}$ to produce a prompt-enhanced feature map \( F_{D_{i+1}} \) as follows:
\begin{equation}
    F_{D_{i+1}} = \operatorname{FFN}\Biggl(\operatorname{L}\Bigl(\underbrace{\operatorname{MCA}\big(\operatorname{L}(F_{D_i}), \operatorname{L}(\bm{\hat{z}}_{i})\big)}_{\operatorname{Cross-Attention} \operatorname{Output}} + F_{D_i}\Bigr)\Biggr) + F_{D_i},
\end{equation}
\noindent where the $F_{D_i}$ serves as the Query vector, and the posterior prompt tokens $\bm{\hat{z}}_{i}$ serve as the Key and Value vectors. By employing this integration mechanism at multiple stages of the student model, we ensure that the model effectively integrates multi-scale contextual information. The posterior tokens from all scales, $\{\bm{\hat{z}}_{i}\}_{i=1}^{M}$, along with the posterior feature vector $\bm{\hat{z}}_{0}$ produced by the anomaly classifier $\zeta$, are concatenated into ${\bm{\hat{z}}}_{cat}$. This concatenated representation is then passed through a global average pooling $\operatorname{GAP}$, followed by a linear layer for the final anomaly classification. The model's output, \(\hat{y}_{i}\), is computed as \(\hat{y}_{i} = \operatorname{Linear}(\operatorname{GAP}(\bm{\hat{z}}_{\text{cat}}))\). The model is then optimized using the cross-entropy loss, defined as \(\mathcal{L}_{{CE}} = -\sum_{i=1}^{N} y_i \log(\hat{y}_{i})\), where \(y_i\) is the ground truth label and \(\hat{y}_{i}\) is the predicted probability for class \(i\). Our hierarchical, context-aware approach facilitates learning and applying class-specific prompt tokens at multiple scales, substantially improving anomaly detection across diverse classes.

\subsection{Domain Adapter}

In the context of unsupervised anomaly detection, the ability to generalize across different domains is crucial, particularly when neglecting distribution shifts that can severely impair model performance. To enhance the robustness of our model to distribution shifts, we propose a novel domain adapter that rectifies diverse styles of the OOD target domain to align with the ID source domain. This alignment is achieved by equipping the residual blocks within the student decoder with AdaIN layers. These layers, characterized by scale ($\gamma$) and shift ($\beta$) parameters, are dynamically controlled by the domain adapter $\xi$. The primary objective of $\xi$ is to align the affine transformation parameters of the AdaIN layers between the OOD target domain and the ID source domain, ensuring consistent feature representation across varying domains.

With no prior knowledge of the target domains, we first synthesize diverse OOD normal data using standard data augmentation. Equipped with the synthesized OOD target domain styles, we then employ a style consistency loss function to regularize the domain adapter $\xi$, ensuring that the synthetic OOD styles closely align with the source domain styles. Specifically, the style consistency loss function is formulated to maximize the cosine similarity between the style codes of the ID source domain and those of the OOD target domain, defined as:
\begin{equation}
\mathcal{L}_{{CS}} = 1 -  \frac{\xi(\mathbf{x}_{\text{ID}})}{\|\xi(\mathbf{x}_{\text{ID}})\|_{2}} \cdot \frac{\xi(\mathbf{x}_{\text{OOD}})}{\|\xi(\mathbf{x}_{\text{OOD}})\|_{2}}, 
\end{equation}
\noindent where $\mathbf{x}_{\text{ID}}$ and $\mathbf{x}_{\text{OOD}}$ denotes the input normal data and the corresponding synthetic normal data from the OOD domain. The style codes are generated by the domain adapter, which is a pre-trained ResNet model \cite{he2016deep}. Maximizing cosine similarity encourages the feature distributions of the OOD domain to closely align with those of the ID domain, promoting greater domain invariance. Once the styles of the OOD target domain are aligned with the source domain, the adapter empowers the well-trained student decoder, optimized using extensive ID source domain data, to effectively process the rectified features on the fly. During testing, the domain adapter operates directly on the features of the OOD target domain, aligning their styles in real time without necessitating back-propagation. This allows the student decoder to adapt to the specific characteristics of each test image dynamically, maintaining robust performance across varying domains. The effectiveness of this approach is encapsulated in the style consistency learning process, which minimizes the divergence between the style codes from different domains. This drives our model to learn domain-invariant representations, thereby enhancing its generalizability and robustness to distribution shifts, which is pivotal for reliable MUAD across diverse domain shifts.

\subsection{Training Objective}%\paragraph{Training Objective.} 
The primary training objective within the RD framework is to minimize the feature discrepancy between the teacher and student networks, specifically for normal images, ensuring precise anomaly detection by enforcing feature alignment. To achieve this, we employ a vector-wise cosine similarity loss, denoted as \( \mathcal{L}_{KD} \), which quantifies the alignment in channel dimension between the feature maps produced by the teacher encoder $E$ and student decoder $D$. The loss function is defined as follows:
\begin{equation}
\mathcal{L}_{KD} = 1 - \sum_{i=1}^{M} \left( \frac{F_{E_i}^\top}{\| F_{E_i} \|_{2}} \cdot \frac{F_{D_i}}{\| F_{D_i} \|_{2}} \right),
\end{equation}
\noindent where \( F_{E_i} \) and \( F_{D_i} \) represent the feature maps of the teacher and student models at the \(i^{\text{th}}\) layer. The term \( \| \cdot \|_2 \) denotes the \( L_2 \)-norm, which is used to normalize the feature vectors. By summing over all \( M \) layers, the loss function integrates similarity across the hierarchical structure of the feature representations, promoting a comprehensive knowledge transfer from the teacher to the student.

Finally, the overall objective function in our training optimization integrates multiple loss functions to achieve both accurate anomaly detection and robustness against domain shifts. The total loss function, \( \mathcal{L}_{total} \), is constructed from the knowledge distillation loss \( \mathcal{L}_{KD} \), cross-entropy loss \( \mathcal{L}_{{CE}} \) for classifying different anomaly categories, and style consistency loss \( L_{{S}} \) to enforce domain invariance. The overall loss function is defined as follows:

\begin{equation}\label{eq.21}
\mathcal{L}_{total} = \eta \mathcal{L}_{KD} + \delta \mathcal{L}_{{CE}} + \mu \mathcal{L}_{{CS}},
\end{equation}

\noindent where \( \eta \), \( \delta \), and \( \mu \) represent the weighting hyperparameters that balance the contributions of different loss functions. During training, the RD framework utilizes normal data to train the student decoder to faithfully reconstruct the multi-scale features of the teacher encoder. This optimization ensures that the student decoder captures the underlying distribution of normal data in different anomaly classes effectively. By integrating classification and style consistency losses into our objective, our approach not only aligns the feature-level representations across different domains with style consistency learning but also leverages class-specific prompt tokens to enhance the overall robustness and performance of MUAD under varying conditions. Having trained our proposed model, during inference, inability of the student decoder to accurately replicate the feature representations from the teacher when confronted with anomalous samples results in a higher loss value, which serves as an indicator of potential anomalies. The anomaly score is then calculated based on the baseline RD framework \cite{guo2024recontrast,tien2023revisiting,deng2022anomaly}. In conclusion, our class-aware prompt integration and the domain adapter significantly enhance the robustness and generalization capabilities of our MUAD framework, particularly under challenging domain shifts.

\begin{table*}[ht]
\centering
\caption{Quantitative comparison with SOTA methods on MVTec-AD \cite{bergmann2019mvtec} benchmark. The results display the performance in anomaly detection and localization, reported as I-AUROC\% / P-AUROC\%. The best results are highlighted in bold.}
\resizebox{\textwidth}{!}{%
\begin{tabular}{l|c|ccccccccccc|c|}
\hline
\multicolumn{2}{c|}{Category/Method}    & PaDiM \cite{defard2021padim} & MKD \cite{salehi2021multiresolution} & DRAEM \cite{zavrtanik2021draem} & PatchCore \cite{roth2022towards} & SimpleNet \cite{liu2023simplenet} & UniAD \cite{Unified2024} & OmniAL \cite{zhao2023omnial} &DiAD \cite{DiAD}& ROADS (Ours)  \\ \hline
\multirow {11}{*}{\rotatebox[origin=c]{90}{Object}}&Bottle      & 97.9 / 96.1 & 98.7 / 91.8 & 97.5 / 87.6 &  \textbf{100} / 98.4 & 86.5 / 88.1 & 99.7 / 98.1 &  \textbf{100} / \textbf{99.2} & {99.7} / {98.4}&  \textbf{100} / 99.1\\

&Cable       & 70.9 / 81.0 & 78.2 / 89.3 &  57.8 / 71.3 & 99.2 / 97.3 &  71.5 / 79.3 &   95.2 / 97.3  & 98.2 / 97.3  &{94.8} / {96.8}&  \textbf{99.3} /  \textbf{99.0}\\ 

&Capsule     & 73.4 / 96.9 & 68.3 / 88.3 & 65.3 / 50.5 & 85.6 / 95.2 & 77.8 / 89.4 & 86.9 / 98.5 &  95.2 / 96.9 & {89.0} / {97.1}&  \textbf{96.0} /  \textbf{99.1}\\ 

&Hazelnut    & 85.5 / 96.3 & 97.1 / 91.2 & 93.7 / 96.9 & 100 / 98.9 & 94.3 / 95.9 & 99.8 / 98.1 & 95.6 / 98.4 & {99.5} / {98.3}&  \textbf{100} /  \textbf{98.9}\\ 

&Metal Nut   & 88.0 / 84.8 & 64.9 / 64.2 &  72.8 / 62.2 & \textbf{99.9} / 98.4 &  87.8 / 87.0 &  99.2 / 94.8 & 99.2 / \textbf{99.1} & {99.1} / {97.3}& 99.7 / 98.2\\ 

&Pill        & 68.8 / 87.7 & 79.7 / 69.7 & 82.2 / 94.4 &  93.3 / 95.7 & 80.2 / 90.7 & 93.7 / 95.0 & \textbf{97.2} / \textbf{98.9}  & {95.7} / {95.7} & 96.2 / 98.0\\ 

&Screw       & 56.9 / 94.1 &  75.6 / 92.1 &  92.0 / 95.5& 82.9 / 95.9 &  72.8 / 85.7&  87.5 / 98.3& 88.0 / 98.0& {90.7} / {97.9}&  \textbf{98.5} /  \textbf{99.6} \\ 

&Toothbrush  & 95.3 / 95.6 &  75.3 / 88.9  & 90.6 / 97.7 &  88.9 / 98.2&87.8 / 96.4 & 94.2 / 98.4 &  \textbf{100} / \textbf{99.4} & {99.7} / {99.0}& 99.2/ 98.7\\

&Transistor  & 86.6 / 92.3 &  73.4 / 71.7 & 74.8 / 64.5 & 96.7 / 89.3 & 79.7 / 83.3 &  \textbf{99.8} /  \textbf{97.9} & 93.8 / 93.3 & \textbf{{99.8}} / {95.1}& 96.3 / 95.8\\ 

&Zipper      & 79.7 / 94.8 &  87.4 / 86.1 & 98.8 / 98.3 & 91.9 / 95.5 & 88.5 / 84.3 & 95.8 / 96.8  &  \textbf{100} /  \textbf{99.5} & {95.1} / {96.2}& 99.6 / 98.5 \\\cline { 2 - 11 }

% &\textit{Average}        &   &  &  &  & &  &  &  \\ \hline

\multirow {6}{*}{\rotatebox[origin=c]{90}{{Texture}}}&Carpet      & 93.8 / 97.6 & 69.8 / 95.5 & 98.0 / 98.6 &  96.1 / 98.7 & 87.6 / 89.5 &  \textbf{99.8} / 98.5 & 98.7 /  \textbf{99.4} & {99.4} / {98.6}& 99.4 / 99.2\\ 

&Grid        & 73.9 / 71.0 & 83.8 / 82.3& 99.3 / 98.7 &  97.1 / 96.6 &  79.1 / 69.9  & 98.2 / 96.5 & 99.9 /  \textbf{99.4} & {98.5} / {96.6}&  \textbf{100} / 99.3 \\ 

&Leather     & 99.9 / 84.8 & 93.6 / 96.7 &  98.7 / 97.3  &  \textbf{100} / 99.4  & 95.2 / 96.6&  \textbf{100} / 98.8 & 99.0 /  {99.3} & {99.8} / {98.8}&  \textbf{100} /  \textbf{99.5}\\ 

&Tile        & 93.3 / 80.5 & 89.5 / 85.3 & 99.8 / 98.0 & 99.9 / 95.7 & 97.9 / 91.6 & 99.3 / 91.8 &  \textbf{99.6} /  \textbf{99.0} & {96.8} / {92.4}& 99.2 / 96.5 \\ 

&Wood        & 98.4 / 89.1  & 93.4 / 80.5 & 99.8 / 96.0 & 98.4 / 93.5& 97.5 / 87.0 & 98.6 / 93.2 & 93.2 /  \textbf{97.4} & {99.7} / {93.3}&  \textbf{99.1} / 96.0 \\ \cline { 1 - 11 }

% &\textit{Average}        &   &  &  &  & &  &  &  \\ \hline

\multicolumn{2}{c|}{\textit{Total Average} }      & 84.2 / 89.5 & 81.9 / 84.9& 88.1 / 87.2 & 95.3 / 96.4 & 85.6 / 87.6 &96.5 / 96.80& 97.1 / {98.3}& {97.2} / {96.8}& \textbf{98.83} / 
 \textbf{98.36 }\\ \hline
\end{tabular}%
}
\label{mvtec_results}

\end{table*}

\vspace{1mm}
\section{Experiments}
\label{sec:Experiments}
\subsection{Experimental settings}

To evaluate the anomaly detection performance of our proposed method, we follow \cite{cao2023anomaly} to create our benchmark and utilize the MVTec-AD \cite{bergmann2019mvtec} and VISA \cite{zou2022spot} datasets for our experiments. The MVTec-AD dataset \cite{bergmann2019mvtec} is the most widely used industrial defect detection dataset, containing a training set $\mathcal{D}_s$ with 3,629 normal images and a test set with 1,725 images, including both defective and non-defective examples. It features 15 subsets for texture and object anomalies. The VISA dataset \cite{zou2022spot} contains 12 objects with 9,621 normal and 1,200 anomalous images, organized into subsets with varying complexity. Anomalies include surface defects (scratches, dents) and structural issues (missing parts). While the original MVTec-AD and VISA datasets do not contain OOD target domains, we introduce domain shifts during testing to simulate such scenarios. Specifically, we follow the benchmark outlined in \cite{cao2023anomaly} to create a target domain \( \mathcal{T} \) by applying four types of visual corruptions (brightness, contrast, defocus blur, and Gaussian noise) based on \cite{hendrycks2018benchmarking}, with no overlap between our data augmentation and these corruptions. The test set \( \mathcal{D}_t\) now includes samples from both the ID source domain \( \mathcal{S} \) and the corrupted OOD target domain \( \mathcal{T} \), with corruption severity set to level 3 for both the MVTec-AD and VISA datasets.

 % to create synthesized OOD normal data

\noindent\textbf{Implementation Detail.}  Our ROADS framework builds upon the RD \cite{deng2022anomaly}, following its settings for image size, optimization, backbone, and anomaly score computation. To train the domain adapter, we use standard augmentations like color adjustments, posterization, and solarization. We compare our approach with several SOTA anomaly detection studies. Our evaluation includes comparisons with MUAD studies such as UniAD \cite{Unified2024}, ViTAD \cite{vitad}, and OmniAL \cite{zhao2023omnial}, alongside leading single-class methods with multi-class training setup such as PatchCore \cite{roth2022towards}, GNL \cite{cao2023anomaly}, and SimpleNet \cite{liu2023simplenet}. In line with previous studies, we assess the effectiveness of anomaly detection and localization using the Area Under the Receiver Operating Characteristic curve at the image-level (I-AUROC) and the pixel-level (P-AUROC). However, since anomalous regions often constitute a small fraction of the image, P-AUROC may not effectively capture localization precision. To address this, we use the Area Under the Per-Region Overlap (P-AUPRO), ensuring consistent sensitivity across different scales.

\subsection{In-distribution Evaluation}

While our main objective is to enhance the robustness of MUAD against OOD distributions, we yield great performance in the ID distributions,  compared to the SOTA studies. In this respect, we evaluate our method on the MVTec-AD dataset against SOTA anomaly detection techniques.

\noindent \textbf{Quantitative Evaluation.} As reported in Table \ref{mvtec_results}, our method demonstrates outstanding performance in both anomaly detection and localization tasks, achieving competitive results on the MVTec-AD dataset. In the context of anomaly detection, using the I-AUROC metric, ROADS outperforms previous methods. Specifically, it surpasses OmniAL \cite{zhao2023omnial} and UniAD \cite{Unified2024} methods, with an average gain of 1.66\% and 2.33\% in terms of I-AUROC score on MVTec-AD dataset. The results for separate models like PaDiM \cite{defard2021padim} and DRAEM \cite{zavrtanik2021draem}, show considerable incapabilities of these models in capturing normal patterns across diverse classes. Moreover, in terms of anomaly localization, our superiority is also maintained over the SOTA studies. These improvements further emphasize the competitive performance of our method in both detecting and localizing anomalies across multiple classes. An explanation for such satisfactory ID anomaly detection performance is attributed to the proposed class-aware prompt integration. This mechanism empowers our student decoder to learn more discriminative boundaries between different classes. Incorporating learnable prompt tokens directly associated with diverse anomaly classes significantly reduces inter-class interference. This approach provides the model with precise contextual information for each anomaly class, which is essential for accurate and effective MUAD.

\noindent\textbf{Qualitative Evaluation.} To visually underscore the effectiveness of our approach, Figure \ref{heat} presents qualitative localization results for both UniAD \cite{Unified2024}  and the proposed ROADS method on the MVTec-AD \cite{bergmann2019mvtec} dataset, evaluated under ID setting. The visualizations clearly demonstrate that our method locates anomalies more accurately in ID setting when compared to UniAD \cite{Unified2024}.

\begin{table*}[ht]
\centering
\caption{Quantitative comparison with SOTA methods on the MVTec-AD \cite{bergmann2019mvtec} benchmark under OOD setting. The results display the performance in anomaly detection and localization, reported as I-AUROC\% / P-AUPRO\%. The best results are highlighted in bold.}
\label{mvtec_OOD}
\resizebox{\textwidth}{!}{%
\begin{tabular}{lc|cccc|cccc|cccc|cccc}
\hline

\multicolumn{2}{c|}{\multirow{2}{*}{Category}} & \multicolumn{4}{c|}{Brightness} & \multicolumn{4}{c|}{Contrast} & \multicolumn{4}{c|}{Blur} & \multicolumn{4}{c}{Gaussian Noise} \\

\cline {3-18} \rule{0pt}{2ex} 
& & UniAD \cite{Unified2024} & GNL \cite{cao2023anomaly} & ViTAD \cite{vitad} & ROADS & UniAD \cite{Unified2024} & GNL \cite{cao2023anomaly} & ViTAD \cite{vitad} & ROADS & UniAD \cite{Unified2024} & GNL \cite{cao2023anomaly} & ViTAD \cite{vitad} & ROADS & UniAD \cite{Unified2024} & GNL \cite{cao2023anomaly} & ViTAD \cite{vitad} & ROADS \\ \hline

\multirow{11}{*}{\rotatebox[origin=c]{90}{Object}}
&Bottle & 99.0/\textbf{95.1} & 98.4/90.8 & \textbf{100}/92.3 & \textbf{100}/95.0 & 98.7/84.1 & 97.5/84.7 & 99.5/90.2 & \textbf{99.8}/\textbf{94.0} & 99.3/94.1 & 99.7/93.5 & 99.9/94.5 & \textbf{100}/\textbf{98.6} & 97.3/82.5 & 69.5/50.9 & {90.4}/60.6 & \textbf{96.9}/\textbf{88.9} \\

&Cable & 95.3/82.4 & 84.7/76.3 & \textbf{98.2}/89.1 & 97.7/88.9 & 89.2/80.0 & 72.1/71.5 & 79.4/80.9 & \textbf{97.5}/\textbf{90.4} & 93.9/83.0 & 82.5/73.8 & \textbf{98.4}/89.1 & 97.9/\textbf{91.5} & 92.7/82.3 & 81.5/56.2 & {91.7}/83.8 & \textbf{98.1}/\textbf{91.2} \\

&Capsule & 72.0/82.7 & 72.5/\textbf{84.9} & 78.0/64.9 & \textbf{86.8}/80.9 & 58.4/67.1 & 74.8/\textbf{84.0} & 71.3/77.0 & \textbf{95.5}/\textbf{87.4} & 70.5/82.9 & 80.0/{88.5} & \textbf{93.6}/88.2 & 90.1/\textbf{89.0} & 67.2/78.7 & 59.5/59.1 & 61.6/59.0 & \textbf{73.4}/\textbf{81.5} \\

&Hazelnut & 99.4/92.2 & 99.4/95.2 & 99.4/93.4 & \textbf{100}/\textbf{96.2} & 98.4/91.2 & \textbf{98.7}/\textbf{94.4} & 98.0/93.1 & 94.1/93.3 & 99.5/92.3 & 99.0/94.5 & 99.9/95.1 & \textbf{100}/\textbf{95.6} & {99.7}/91.0 & 76.1/43.7 & 76.4/48.3 & \textbf{99.9}/\textbf{92.8} \\

&Metal Nut & 97.4/{77.9} & 98.2/74.6 & 99.1/83.1 & \textbf{99.9}/\textbf{92.5} & 97.7/75.7 & 96.5/82.4 & 97.8/86.4 & \textbf{99.9}/\textbf{90.8} & 97.4/79.0 & 98.2/89.8 & 99.6/\textbf{92.2} & \textbf{99.8}/91.6 & \textbf{97.0}/67.6 & 64.7/38.5 & 90.9/28.5 & 95.3/\textbf{81.6} \\

&Pill & 70.4/84.8 & 82.6/81.9 & 85.5/82.9 & \textbf{88.2}/\textbf{90.2} & 59.0/80.6 & 82.4/{89.5} & 80.2/81.8 & \textbf{95.6}/\textbf{93.5} & 72.8/87.1 & 92.4/92.0 & \textbf{97.0}/\textbf{94.8} & 96.0/\textbf{94.8} & 58.4/\textbf{82.6 }& 57.1/47.4 & 59.6/29.2 & \textbf{75.4}/{82.3} \\

&Screw & {88.2}/84.7 & 63.1/\textbf{93.4} & 87.9/42.5 & \textbf{90.1}/62.2 & 78.8/89.6 & 84.9/\textbf{95.0} & 83.4/90.1 & \textbf{90.5}/{94.0} & {87.9}/88.8 & 86.2/{95.6} & 87.4/92.5 & \textbf{91.3}/\textbf{96.9} & \textbf{85.4}/90.5 & 43.9/69.9 & 14.2/17.1 & 75.4/\textbf{91.1} \\

&Toothbrush & 78.3/81.5 & 92.2/\textbf{92.6} & 90.2/88.3 & \textbf{94.1}/90.3 & 97.2/80.0 & {95.8}/\textbf{89.7} & \textbf{98.0}/86.0 & 91.4/89.3 & 81.9/86.2 & {99.1}/\textbf{92.6} & 99.6/90.2 & \textbf{99.7}/90.1 & {93.6}/80.0 & 84.4/58.9 & 79.7/35.3 & \textbf{99.7}/\textbf{87.7} \\

&Transistor & \textbf{99.4}/91.8 & 88.6/{92.6} & 92.2/69.5 & 98.7/\textbf{94.4} & 93.3/84.6 & 83.2/89.7 & 79.2/60.7 & \textbf{99.0}/\textbf{93.8} & {99.3}/92.5 & 90.7/68.9 & 97.6/76.0 & \textbf{99.4}/\textbf{97.3} & \textbf{97.3}/90.4 & 81.4/53.2 & 81.0/53.8 & \textbf{97.3}/\textbf{92.5} \\

&Zipper & 87.7/87.5 & 95.0/{91.7} & 98.8/88.9 & \textbf{99.6}/\textbf{93.5} & 71.2/36.4 & 93.5/75.9 & 98.2/{90.9} & \textbf{99.0}/\textbf{91.0} & 88.1/87.5 & 96.8/{89.0} & 97.2/84.0 & \textbf{99.7}/\textbf{90.5} & {82.4}/56.1 & 47.9/15.7 & 82.0/36.0 & \textbf{93.0}/\textbf{76.6}\\ \cline {2-18}

\multirow{6}{*}{\rotatebox[origin=c]{90}{Texture}}
&Carpet & \textbf{99.7}/\textbf{95.3} & 95.2/92.5 & 99.3/93.9 & 97.0/94.5 & \textbf{99.4}/\textbf{94.3} & 94.1/85.0 & 98.2/91.8 & 98.9/93.8 & \textbf{99.6}/94.1 & 96.1/88.7 & 99.7/95.2 & 98.0/\textbf{98.9} & \textbf{99.9}/93.8 & 96.8/\textbf{94.5} & 96.4/92.4 & 97.7/93.8 \\

&Grid & 98.3/85.5 & 96.8/\textbf{96.8} & {99.5}/96.0 & \textbf{99.6}/96.6 & 98.3/85.9 & 94.5/\textbf{96.4} & 98.4/92.9 & \textbf{99.4}/94.9 & {99.2}/89.3 & 95.9/{97.0} & 99.0/95.1 & \textbf{99.4}/\textbf{98.8} & {96.4}/87.9 & 82.9/93.4 & 89.2/89.4 & \textbf{98.7}/\textbf{95.4} \\

&Leather & \textbf{100}/98.3 & 99.7/\textbf{98.5} & \textbf{100}/97.4 & \textbf{100}/98.0 & \textbf{100}/\textbf{97.8} & 97.1/98.0 & 99.2/95.1 & \textbf{100}/96.8 & \textbf{100}/\textbf{98.2} & 99.8/97.9 & \textbf{100}/97.2 & 99.8/97.2 & \textbf{100}/\textbf{97.7} & 54.1/41.6 & \textbf{100}/95.6 & 99.7/95.9 \\

&Tile & 94.7/75.0 & 97.6/84.3 & \textbf{100}/\textbf{87.5} & \textbf{100}/86.1 & {95.6}/72.1 & 91.0/61.2 & 96.1/78.9 & \textbf{100}/\textbf{83.8} & {96.4}/71.2 & 93.6/66.9 & \textbf{100}/\textbf{85.3} & 99.9/81.3 & 91.9/{74.4} & 90.3/37.2 & \textbf{100}/\textbf{85.3} & 99.6/{82.5} \\

&Wood & 95.1/84.7 & {96.4}/\textbf{90.6} & 96.7/81.7 & \textbf{97.4}/87.1 & \textbf{98.4}/83.6 & 96.4/\textbf{87.1} & 87.9/65.5 & 96.3/84.1 & {96.9}/87.3 & 96.2/87.9 & \textbf{97.4}/87.8 & 97.0/\textbf{88.3} & {96.7}/83.6 & 76.7/44.9 & {96.3}/76.6 & \textbf{98.2}/\textbf{83.7} \\ \cline {1-18}

\multicolumn{2}{c|}{\textit{Total Average}} & 91.7/86.6 & 90.6/{89.1} & {95.0}/83.4 & \textbf{96.0}/\textbf{89.7} & 88.9/80.2 & 90.1/85.6 & 91.0/84.1 & \textbf{97.1}/\textbf{91.4} & 92.2/87.6 & 93.7/87.7 & {97.7}/90.5 & \textbf{97.8}/\textbf{93.3} & {90.4}/82.6 & 71.1/53.6 & 80.6/59.4 & \textbf{93.2}/\textbf{87.8} \\ \hline

\end{tabular}%
}
\end{table*}

\subsection{Out-of-Distribution Evaluation}

This subsection aims to assess the effectiveness of ROADS in addressing distribution shift challenges against SOTA methods, including UniAD \cite{Unified2024}, GNL \cite{cao2023anomaly}, ViTAD \cite{vitad}, DiAD \cite{DiAD}, RD++ \cite{tien2023revisiting}, and SimpleNet \cite{liu2023simplenet}.

\noindent\textbf{Quantitative Evaluation.} The performance of ROADS is benchmarked on the MVTec-AD and VISA datasets, both of which introduce various OOD conditions, including brightness, contrast, blur, and Gaussian noise perturbations. The results in Table \ref{mvtec_OOD} clearly demonstrate that ROADS consistently outperforms these methods across evaluated scenarios.
On the MVTec-AD dataset, ROADS exhibits exceptional robustness in detecting and localizing anomalies across all object and texture categories under OOD conditions, as shown in Table \ref{mvtec_OOD}. For instance, in the presence of Gaussian noise, ROADS achieves an average I-AUROC of 93.2\%, significantly higher than UniAD's 90.04\% and GNL's 71.1\%. Under contrast variation, ROADS maintains high detection accuracy with an I-AUROC of 97.1\%, outperforming ViTAD by 6.1\%. We also secure first place in anomaly localization with a wide margin in OOD scenarios. These results underscore the ability of ROADS to effectively handle OOD challenges, maintaining robustness where other methods experience performance degradation.

Similarly, on the VISA dataset, as detailed in Table \ref{VISA_OOD}, which features more complex structures and multiple instance anomalies, ROADS surpasses all compared methods in both  I-AUROC and P-AUPRO metrics under all OOD conditions. For example, in scenarios involving blur degradation, where image details become obscured, ROADS maintains an impressive P-AUPRO of 89.55\%, compared to 72.0\% for SimpleNet and 81.62\% for ViTAD. This level of robustness is particularly crucial in industrial settings, where maintaining consistent detection accuracy is essential despite unpredictable distribution shifts.

The superior performance of ROADS can be attributed to the domain adapter. The domain adapter in ROADS plays a pivotal role in aligning feature representations between OOD target domains and ID source domains. The domain adapter dynamically adjusts the AdaIN parameters to different domains, ensuring consistent and robust performance even under significant distribution shifts.  The superiority of our method over approaches like UniAD \cite{Unified2024}, ViTAD \cite{vitad}, SimpleNet \cite{liu2023simplenet}, and RD++ \cite{tien2023revisiting} lies in its ability to effectively handle OOD conditions, where these methods struggle due to the inter-class interference and the lack of domain alignment mechanisms. Even the GNL method \cite{cao2023anomaly}, which is specifically designed for OOD settings in anomaly detection, falls short in multi-class scenarios because of its reliance on static feature representations. Overall, the integration of the domain adapter enables ROADS to maintain high detection accuracy across challenging OOD conditions. 

\begin{table}[t]
\centering
\caption{Quantitative comparison with SOTA methods on benchmark VISA \cite{zou2022spot} under OOD and ID settings. Results for anomaly detection and localization are shown as I-AUROC\% / P-AUPRO\%. The best results are highlighted in bold.}
\label{VISA_OOD}
\resizebox{0.48\textwidth}{!}{%
\begin{tabular}{l|ccccc}
\hline

  {{Category}}& ID & Brightness & Contrast & Blur &Gaussian Noise  \\ \hline

UniAD \cite{Unified2024}& 90.33 / 86.99 & 81.19 / 80.87 & 78.16 / 78.27 & 90.61 / 85.88 & 85.68 / 81.82 \\

ViTAD \cite{vitad}& 90.58 / 84.77 & 68.73 / 61.55 & 77.79 / 70.98 & 89.92 / 81.62 & 75.53 / 50.54 \\

DiAD \cite{DiAD}& 90.52 / 44.36 & 76.14 / 32.95 & 72.41 / 29.56 & 88.23 / 41.03 & 83.48 / 37.89 \\

RD++ \cite{tien2023revisiting}& 93.94 / 91.79 & 73.89 / 75.16 & 81.92 / 84.36 & 92.58 / 88.52 & 75.19 / 76.56 \\

SimpleNet \cite{liu2023simplenet}& 87.94 / 82.66  & 61.35 / 46.34 & 56.51 / 55.91 &  79.28 / 72.00 & 61.52 / 42.93 \\

ROADS (Ours)& \textbf{95.42} / \textbf{92.27}&  \textbf{85.98} / \textbf{78.07} & \textbf{83.88 }/ \textbf{85.27} & \textbf{94.62} / \textbf{89.55} & \textbf{88.63} / \textbf{84.08 }\\ 

\hline
\end{tabular}%
}
\label{Table1}%

\end{table}

\begin{table}
\center
\caption{Ablation study on the key components of ROADS and the weighting hyperparameters that adjust the impact of different loss functions. Anomaly detection  results are reported as P-AUPRO\%.}
\resizebox{0.3\textwidth}{!}{%
\begin{tabular}{l|ccc|c|c}

\hline  \multirow{2}{*}{ Model } &\multicolumn{3}{c|}{ Components } &    \multirow{2}{*}{ID}&   \multirow{2}{*}{OOD}\\

 \cline { 2 - 4}&$\mathcal{L}_{{CS}}$ & $\mathcal{L}_{{CE}}$ & $\mathcal{L}_{KD}$    &  \\
\hline 
ROADS-0 &- & -& $\checkmark$   & 91.15& 82.20\\
ROADS-1 &$\checkmark$& - &  $\checkmark$ &  91.23 & 86.49 \\

ROADS-2 &-&  $\checkmark$&  $\checkmark$ & 93.74  & 85.03\\
ROADS-3 &$\checkmark$&  $\checkmark$& $\checkmark$  &93.69& 90.55\\

\hline  \multirow{2}{*}{ Model } &\multicolumn{3}{c|}{ Hyperparameter} &   \multirow{2}{*}{ID}&   \multirow{2}{*}{OOD}\\

 \cline { 2 - 4}&  $\mu$ & $\delta$ &   $\eta$  &  \\ \hline 

      ROADS-4 & 0.025& 0.025& 0.95  &93.69&  90.55 \\
      ROADS-5 & 0.04&0.01 & 0.95  &93.37& 90.34\\
      ROADS-6 & 0.01&0.04 & 0.95 &92.94&89.72 \\
      ROADS-7 & 0.05& 0.05 & 0.9  &92.61& 89.43\\

\hline
\end{tabular}}
\label{ablation}%
\end{table}%

\begin{figure}[t] 
\centering
    \includegraphics[scale = 0.154]{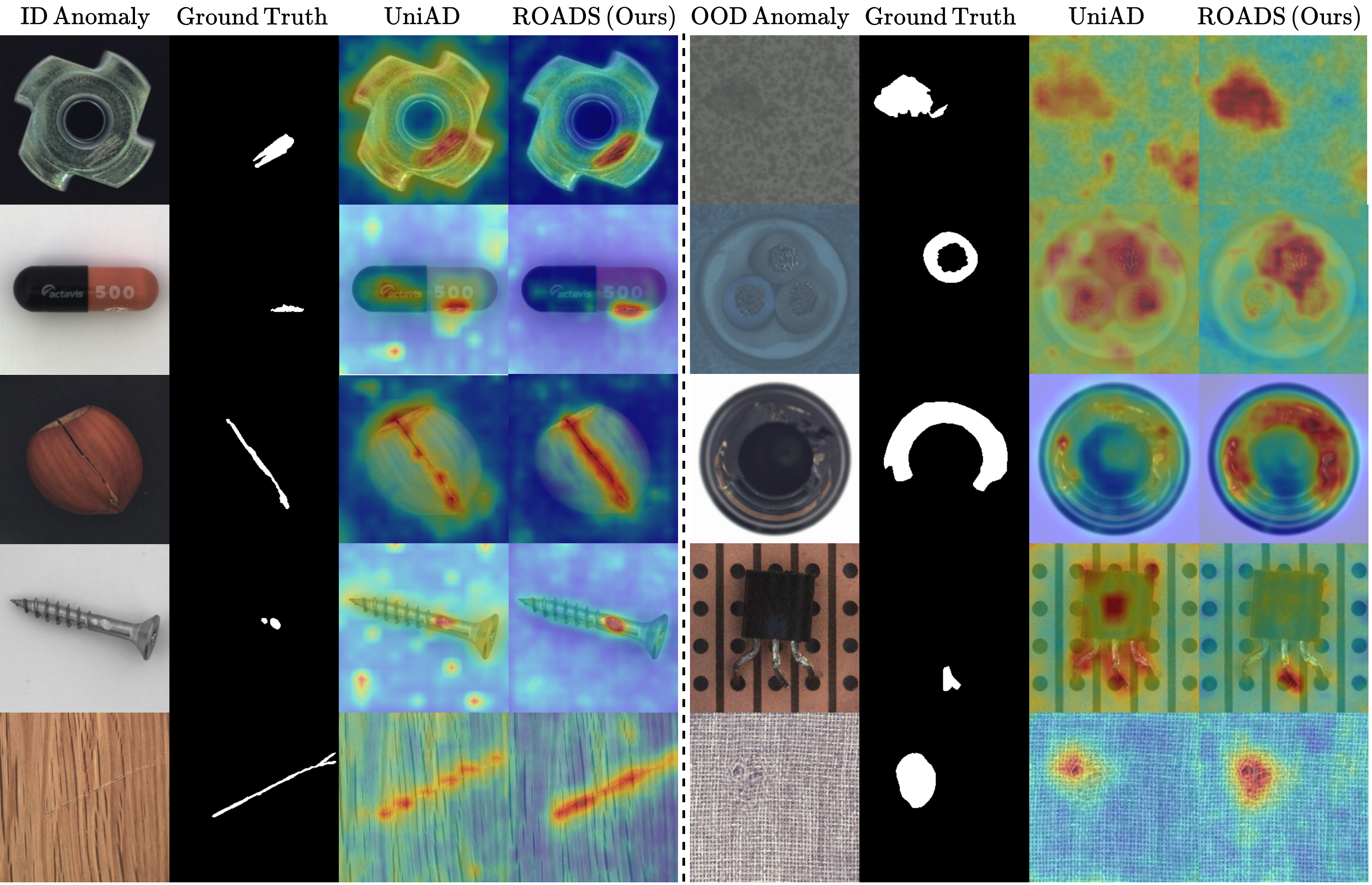}%width=0.5\textwidth mask5 not good 
    \caption{Qualitative comparison between the proposed ROADS method and UniAD \cite{Unified2024} on the MVTec-AD dataset \cite{bergmann2019mvtec} under both ID and OOD settings.}
    \label{heat}
\end{figure}

\noindent\textbf{Qualitative Evaluation.} Figure \ref{heat} illustrates the clear superiority of our method over UniAD \cite{Unified2024} in OOD scenarios. While UniAD \cite{Unified2024} struggles to accurately localize anomalies under distribution shifts, ROADS consistently identifies anomalies, demonstrating robustness in such conditions.

\subsection{Ablation Study} 
We conduct ablation experiments on the MVTec-AD dataset to assess key components of our framework in both ID and OOD scenarios. The average P-AUPRO scores for four OOD conditions (brightness, contrast, defocus blur, and Gaussian noise) are reported for the OOD scenario. Also, we analyze the impact of the weighting hyperparameters on the overall loss function.
% To comprehensively assess the contributions of key components in our proposed framework, we conduct a series of ablation experiments on the MVTec-AD dataset, focusing on both ID and OOD scenarios. The average P-AUPRO\% scores for four OOD conditions (brightness, contrast, defocus blur, and Gaussian noise) are considered for the OOD scenario. In addition, we analyze the impact of the weighting hyperparameters in the overall loss function.

\noindent\textbf{Effectiveness of Class-Aware Prompt Integration.} The class-aware prompt integration mechanism mitigates inter-class interference by incorporating class-specific tokens during anomaly detection. To evaluate its impact, we train a variant of our model without this mechanism, denoted as ROADS-1. The ROADS-3 model, with all components activated, represents the best performance. Removing class-specific prompts in ROADS-1 causes a significant performance drop, with a 2.46\% decrease in the ID setting and 4.06\% in the OOD setting, as measured by P-AUPRO on the MVTec-AD dataset. This emphasizes the role of class-specific prompts as explicit constraints that guide the detector to capture the diversity among anomaly classes.

\noindent\textbf{Effectiveness of Domain Adapter.}
The domain adapter enhances domain invariance by aligning the style codes of ID and OOD domains. Ablation experiment disabling the domain adapter and its $\mathcal{L}_{{CS}}$ loss in ROADS-2 shows a significant performance drop, particularly under severe OOD scenarios, with an average P-AUPRO decrease of 5.52\%. This result highlights its importance in handling domain shifts.

\noindent\textbf{Hyperparameter Analysis in the Loss Function.}  The hyperparameters \( \eta \), \( \delta \), and \( \mu \) play critical roles in balancing knowledge distillation, anomaly classification, and style consistency losses. Increasing \( \eta \) from 0.9 to 0.95 improves anomaly detection, indicating that prioritizing $\mathcal{L}_{KD}$ is more critical than other regularizations. Optimal performance is achieved with \( \delta \) at 0.025, as reducing it to 0.01 leads to lower localization accuracy, suggesting insufficient supervision from class-specific prompt tokens. The style consistency loss weight, \( \mu \), also performs best at 0.025, boosting generalization in OOD scenarios.

\section{Conclusion}

% Also, the results in ID scenarios demonstrate that ROADS not only meets but exceeds the performance of current SOTA methods, making it a significant contribution to the field of MUAD.

This paper presents a robust prompt-driven multi-class anomaly detection framework (ROADS) to address domain shifts in multi-class settings. By integrating a class-aware prompt mechanism and a domain adapter, ROADS effectively addresses the challenges of inter-class interference and domain shifts. The class-specific prompts guide the student decoder, reducing overlap between anomaly classes, while the domain adapter ensures consistency across varying domains by learning domain-invariant representations. Extensive experiments on different datasets demonstrate the superior performance of our method in both anomaly detection and localization, particularly under OOD conditions. 
\textbf{Acknowledgements.} This material is based upon work supported by the National Science Foundation under Grant Numbers CNS-2232048, and CNS-2204445.

%%%%%%%%% REFERENCES
{\small
\bibliographystyle{ieee_fullname}
\bibliography{egbib}
}

\end{document}